%% file: main__7_.tex
\documentclass[11pt,twocolumn]{article}

\usepackage[margin=1in]{geometry}
\usepackage{amsmath}
\usepackage{graphicx}
\usepackage{booktabs}
\usepackage{hyperref}
\usepackage{float}
\usepackage{standalone}
\usepackage{enumitem}

\usepackage{tikz}
\usetikzlibrary{positioning, arrows.meta, calc, decorations.pathreplacing}
\title{Inference-Path Optimization via Circuit Duplication in Frozen Visual Transformers for Marine Species Classification

}

\author{Thomas Manuel Rost}
\date{\today}

\begin{document}

\twocolumn[
\maketitle
\begin{abstract}
Automated underwater species classification is constrained by annotation cost
and environmental variation that limits the transferability of fully supervised
models. Recent work has shown that frozen embeddings from self-supervised vision
foundation models already provide a strong label-efficient baseline for marine
image classification. Here we investigate whether this frozen-embedding regime
can be improved at inference time, without fine-tuning or changing model weights.

We apply Circuit Duplication, an inference-time method originally proposed for
Large Language Models, in which a selected range of transformer layers is
traversed twice during the forward pass. We evaluate on the class-imbalanced
AQUA20 benchmark using frozen DINOv3 embeddings under two settings: global
circuit selection, where a single duplicated circuit is chosen for the full
dataset, and class-specific circuit selection, where each species may receive
a different optimal circuit. Both settings use simple semi-supervised downstream
classifiers.

Circuit Duplication consistently improves over the standard frozen forward pass.
At the maximum label budget, class-specific selection reaches a macro F1 of
0.875, closing the gap to the fully supervised ConvNeXt benchmark (0.889) to
1.4 points without any gradient-based training. Four species exceed their fully
supervised reference, with octopus improving by +12.1 F1 points. Across all
budgets, roughly 75\% of classes prefer a class-specific circuit, indicating
a genuinely class-dependent benefit. To our knowledge, this is the first
application of Circuit Duplication to computer vision.
\end{abstract}
\vspace{1em}
]

\clearpage
\section{Introduction}

Visual classification plays an increasingly important role in marine science \cite{fisher2016fish4knowledge,goodwin2022unlocking,radeta2022deeplearning_oceans}. Underwater imaging is used for biodiversity monitoring, ecological surveys, fisheries assessment, and exploratory observation in environments that are difficult to sample directly \cite{dominguezcarrio2021azordriftcam,goodwin2022unlocking}. However, converting underwater imagery into reliable species-level observations remains challenging. Taxonomic annotation is expensive, requires expertise, and is often difficult to scale \cite{saleh2022survey,mittal2023underwater_survey}. In addition, underwater imagery is affected by turbidity, lighting variation, colour distortion, background complexity, and camera-specific artefacts, all of which complicate supervised learning and reduce transfer across settings \cite{fuad2026aqua20,saleh2022survey}.

Recent progress in self-supervised foundation models has opened a different route \cite{caron2021dino,oquab2023dinov2,dinov32025}. Rather than training a model end-to-end on a domain-specific labeled dataset, one can extract embeddings from a large pre-trained transformer-based image model and use those frozen representations for downstream classification \cite{dosovitskiy2021vit,oquab2023dinov2}. Prior work has shown that this approach can be surprisingly effective in ecological classification tasks \cite{markoff2026vit_clustering,gustineli2024plantclef,picon2026agricultural}, and recent results on the AQUA20 benchmark indicate that frozen DINOv3 embeddings paired with simple semi-supervised methods already provide a strong label-efficient baseline for underwater species recognition \cite{rost2026labelefficientunderwaterspeciesclassification}.

In this paper, we ask whether that frozen-embedding regime can itself be improved at inference time, without fine-tuning and without changing model weights. This is important because it opens a route toward better task-specific representations without requiring the curation of large fully supervised training datasets. Instead of collecting and annotating massive image corpora for end-to-end training, one can begin from a strong frozen foundation model, use only a small number of labeled examples, and optimize the embedding process itself. In addition, computing forward passes for the selection of optimal circuit parameters is much cheaper than fully supervised training pipelines that require significant compute. In a marine setting, where labeled data are costly and often scarce for rare classes, this has clear potential for efficient data collection, rapid deployment, and active-learning workflows.

We study this through Circuit Duplication, a method recently introduced in the context of Large Language Models \cite{ng2026rys}. Rather than changing parameters, Circuit Duplication changes the path taken through the transformer at inference time by duplicating a selected range of layers. The hidden state is re-entered into an earlier point in the stack, so that part of the computation is repeated before the forward pass continues.

Our central question is whether such inference-path modification can improve downstream classification in a frozen foundation-model setting. We study this in two forms. In the first experiment, we search for a single globally optimal duplicated circuit for the dataset as a whole. In the second, we allow the optimal circuit to differ by class and evaluate class-specific circuit selection. The motivation for the second experiment is that marine species differ substantially in morphology, texture, silhouette, and visual context, so the same inference path may not be equally useful for all of them. This lets us test both whether circuit duplication is useful in general, and whether its benefits are partly class-dependent.

The contributions of this paper are threefold. First, to our knowledge, this is the first application of Circuit Duplication to computer vision. Second, it is the first application of the method in marine species classification. Third, it introduces and evaluates two optimization settings, global circuit selection and class-specific circuit selection, showing that inference-time circuit modification consistently improves frozen visual embeddings for downstream classification, that such gains are genuinely class-dependent, and that the resulting system approaches fully supervised performance without any gradient-based training.

\section{Related Work}

\textbf{Label-efficient marine species classification.}
Automated underwater species recognition has historically relied on fully supervised computer vision models trained on labeled marine datasets \cite{fuad2026aqua20}. While such models can achieve strong performance when sufficient labels are available, their practical use is often limited by annotation cost, domain shift, and the difficulty of collecting representative underwater data across sites and conditions.

The AQUA20 benchmark provides an important recent reference point for this problem. It is a challenging underwater species classification dataset with substantial class imbalance and realistic image degradation \cite{fuad2026aqua20}, making it a suitable testbed for label-efficient alternatives to fully supervised training.

\textbf{Frozen foundation model embeddings.}
A major recent development in computer vision has been the emergence of self-supervised vision foundation models such as DINOv2 and DINOv3 \cite{oquab2023dinov2, dinov32025}. These models are trained on very large image corpora without manual labels and produce general-purpose visual representations that transfer well to downstream tasks. Instead of learning a classifier end-to-end, one can use the foundation model as a frozen feature extractor and apply lightweight downstream methods.

Recent work has shown that frozen transformer-based image embeddings can support strong performance in ecological settings, including unsupervised and weakly supervised species-level tasks \cite{markoff2026vit_clustering,rost2026labelefficientunderwaterspeciesclassification}. In the marine domain, this approach is particularly attractive because it shifts most of the representation-learning burden away from scarce labeled datasets.

\textbf{Inference-time architectural modification.}
Most efforts to adapt foundation models to downstream tasks still rely on training. This includes full fine-tuning and parameter-efficient methods such as Low-Rank Adaptation, meaning lightweight trainable additions that modify model behaviour with fewer updated parameters \cite{relaxed_recursive2024}. By contrast, inference-time architectural modification changes the computation performed by the model during the forward pass while leaving all weights fixed.

Circuit Duplication, introduced in the context of Large Language Models \cite{ng2026rys}, is one such method. It duplicates a selected subset of transformer layers by routing the hidden state back through an earlier segment of the network. To our knowledge, this idea has not yet been applied to computer vision or to marine image classification.

\textbf{Gap addressed by this paper.}
Current label-efficient pipelines in marine image classification largely assume a fixed frozen feature extractor. The representation may be strong, but the inference path itself remains static. This paper examines whether that assumption can be relaxed. Specifically, we test whether a frozen visual foundation model can be made more useful for downstream marine classification through inference-path modification, without retraining, and whether such modification operates differently across classes.

\section{Methods}

\subsection{Overview}

Our starting point is a frozen DINOv3 backbone. Rather than using only the standard forward pass, we generate alternative embeddings by modifying the inference path through Circuit Duplication. Each duplicated circuit is defined by a pair of layer indices, and each configuration produces a distinct embedding space. We then evaluate downstream classifiers on those embeddings under two optimization settings: one global and one class-specific.

\subsection{Circuit Duplication in a visual transformer}

\begin{figure*}[t]
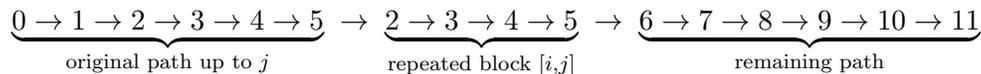

\centering
\[
\underbrace{0 \to 1 \to 2 \to 3 \to 4 \to 5}_{\text{original path up to } j}
\;\to\;
\underbrace{2 \to 3 \to 4 \to 5}_{\text{repeated block } [i,j]}
\;\to\;
\underbrace{6 \to 7 \to 8 \to 9 \to 10 \to 11}_{\text{remaining path}}
\]
\caption{Example of the effective path when layers $i$ through $j$ are repeated.}
\label{fig:effective_path_example}
\end{figure*}

The backbone is a 12-layer DINOv3 Base model with patch size 16 \cite{dinov32025}. In the standard model, the hidden state passes through the transformer blocks once, in order, from layer 0 to layer 11. We define a circuit by two integers, $(i, j)$, with $0 \le i < j < 12$. The model first processes layers 0 through $j$, then re-enters at layer $i$ and repeats layers $i$ through $j$, and finally continues forward through the remaining layers. 
As illustrated in Figure~\ref{fig:effective_path_example}, the duplicated circuit is inserted immediately after its first traversal and before execution continues with subsequent layers.

importantly, no model weights are changed. The only difference is the computational path taken during inference. The duplicated layers are already part of the model, so the method does not require loading additional parameters into memory. As noted in the original RYS work \cite{ng2026rys}, the main cost is additional computation from traversing part of the existing stack more than once. It therefore increases inference time for the duplicated block, but does not require gradient computation, backpropagation, retraining, or additional model-memory overhead. As in the original work \cite{ng2026rys}, the mechanism by which this improves performance remains unclear and is in some sense counter-intuitive. A plausible hypothesis is that transformer residual connections make this kind of re-entrant computation possible without catastrophic degradation, but at present this remains speculative.

We evaluate all 66 possible circuits in a 12-layer model, corresponding to all ordered pairs $(i, j)$ with $i < j$. The standard non-duplicated forward pass is included as both a reference condition and in the case of class based optimsation, as potential optimisation choice.

\subsection{Image preprocessing and embedding extraction}

All images are resized to $518 \times 518$ pixels and passed through the modified DINOv3 backbone. We extract a single image representation by averaging the spatial patch tokens from the final layer output, excluding the classification token and register tokens, and then applying $L_2$ normalization, meaning scaling each vector to unit length.

Each circuit configuration produces its own embedding space. Because these spaces differ, we treat them as separate manifolds rather than as interchangeable feature variants.

\subsection{Dimensionality reduction}

For each circuit configuration, we reduce the 768-dimensional embeddings to 128 dimensions using Principal Component Analysis, a standard linear dimensionality-reduction method that captures the major directions of variance in the data. A separate Principal Component Analysis model is fitted for each circuit. This avoids mixing the geometry of different duplicated inference paths into a shared projection.

As in the baseline study, dimensionality reduction is performed in a transductive manner using the combined train and test embedding set for the relevant circuit configuration. This does not expose test labels, but does allow the geometry of the full sample to shape the representation.

\subsection{Downstream classifiers}

We evaluate several downstream classifiers on the resulting circuit-specific embedding spaces. The methods have been deliberately chosen to be simple in order to show the power of the underlying embedding generation:

\begin{itemize}[leftmargin=1.2em]
    \item \textbf{Label Spreading:} a graph-based semi-supervised method in which labels are diffused over a similarity graph connecting labeled and unlabeled samples \cite{zhu2002label}.
    \item \textbf{Self-Training K-nearest neighbours:} a nearest-neighbour classifier combined with iterative pseudo-labeling \cite{lee2013pseudolabel,yarowsky1995}.
    \item \textbf{Self-Training Support Vector Machine:} a pseudo-labeling procedure built around a Support Vector Machine classifier \cite{lee2013pseudolabel,yarowsky1995}.
    \item \textbf{Seeded K-means:} a centroid-based baseline using labeled seeds for initialization.
    \item \textbf{K-nearest neighbours baseline:} a purely supervised nearest-neighbour classifier trained only on the labeled seeds.
\end{itemize}

These are the same downstream methods used in the baseline study \cite{rost2026labelefficientunderwaterspeciesclassification}, allowing direct comparison between the standard frozen-embedding setting and the circuit-duplicated setting.

\section{Experimental Design}

\subsection{Dataset}

We evaluate on AQUA20, a benchmark dataset for underwater species classification comprising 20 marine categories and an official train-test split \cite{fuad2026aqua20}. The dataset is strongly class-imbalanced and includes challenging visual conditions typical of underwater imagery. This makes it a suitable test case for label-efficient classification methods and for circuit-level optimization.
Conventiently, the authors ot AQUA20 provide the performance of fully supervised state of the art training regiment that we use as benchmark for the performance of our approaches.

\subsection{Three-pool evaluation design}

To avoid leakage, we adopt a three-pool design. First, a validation pool is carved out from the training split before any labeled seeds are assigned. Second, labeled seeds are drawn only from the remaining training subset. Third, the official test set is reserved exclusively for final evaluation.

It is worth noting that the validation pool (40\% of the training split) is
reserved entirely for circuit selection and is never used as labeled data for
the downstream classifiers. Thus, even at the maximum label budget
(denoted 100\% in our experiments), the classifiers are trained on
approximately 60\% of the original training data. The fully supervised
ConvNeXt benchmark, by contrast, is trained on the entire training split. Comparissons therefore are biased against our approach which might potentially perform even better than reported below. 

\subsection{Global circuit optimization}

In the first experiment, we evaluate all 66 possible duplicated circuits together with the standard non-duplicated forward pass and select a single circuit for the dataset as a whole. Circuit quality is measured on the validation pool, allowing comparison of global duplicated inference paths against the standard frozen baseline under the same downstream classifiers. This experiment tests whether circuit duplication is useful as a general architectural intervention even without class-specific adaptation.

\begin{figure*}[t]
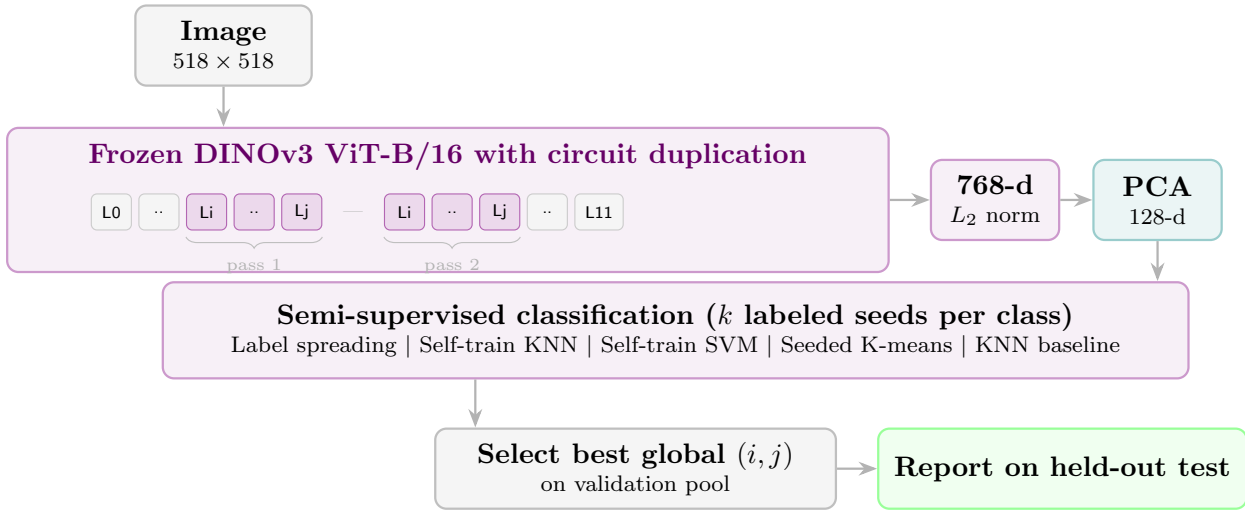

\centering
\includestandalone[width=\textwidth]{exp2_pipeline}
\caption{Pipeline for global circuit selection. All 66 duplicated circuits are swept over the frozen DINOv3 backbone. A single best $(i,j)$ pair is selected on the validation pool across all classes, and final performance is reported on the held-out test set.}
\label{fig:exp2_pipeline}
\end{figure*}

Figure~\ref{fig:exp2_pipeline} summarizes the global circuit selection pipeline.

\subsection{Class-specific circuit optimization}

Visual classification tasks often involve classes with a vide variety of forms, shapes and patterns. Some may be separable primarily through coarse shape, while others may depend more strongly on repeated texture, local structure, or finer-grained visual refinement. If circuit duplication changes the balance of these representational properties, then different classes may benefit from different duplicated regions of the transformer.

For this reason, we evaluate a second optimization strategy. In the class-specific setting, the best-performing circuit is selected independently for each class based on validation set performance. This produces a class-specific mapping from species to circuit configuration via the chosen classificaion method. In other words, the embedding extraction process itself becomes class-dependent, allowing us to compare different circuits on performance for each class in a one-vs-all approach.

Results are averaged across repeated runs with fixed random seeds for each circuit and classifier combination. The key quantity of interest in this experiment is the best-performing combination of circuit and downstream method for each class, allowing us to determine whether some classes benefit systematically from different inference paths than others.

\begin{figure*}[t]
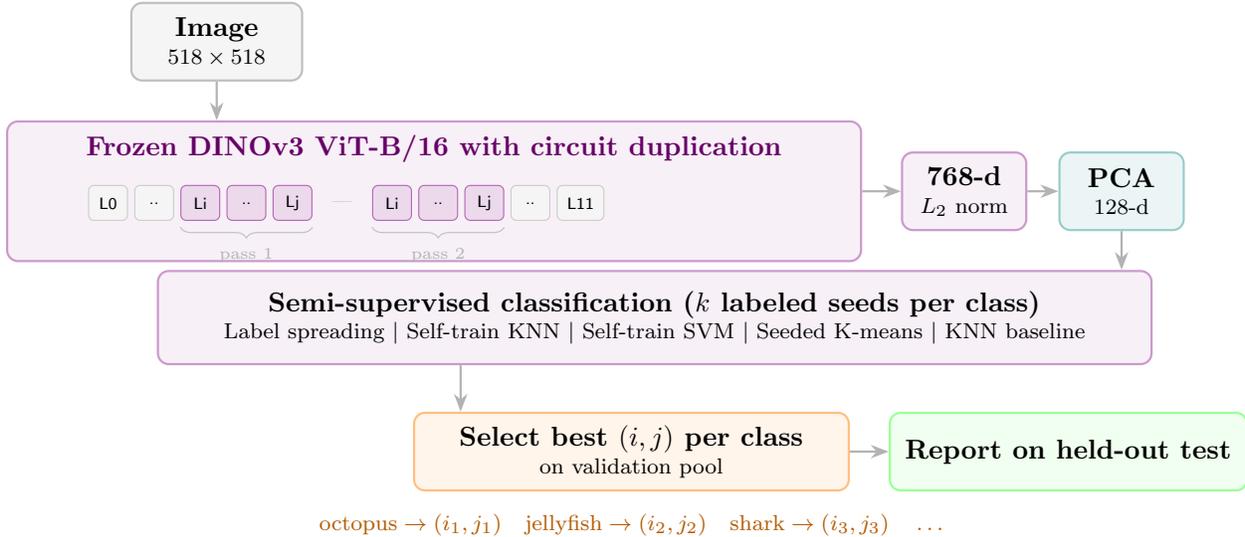

\centering
\includestandalone[width=\textwidth]{exp3_pipeline}
\caption{Pipeline for class-specific circuit selection. The same sweep is performed, but the best $(i,j)$ pair is selected independently for each class on the validation pool. Different species may therefore receive different optimal inference paths through the frozen transformer.}
\label{fig:exp3_pipeline}
\end{figure*}

Figure~\ref{fig:exp3_pipeline} summarizes the class-specific circuit selection pipeline.

\subsection{Label budgets}

We evaluate both absolute and fractional label budgets. Absolute budgets (5, 10, and 20 labeled seeds per class) allow direct control over annotation effort. Fractional budgets (5\%, 10\%, 15\%, and 100\% of the training split) allow comparison with the full-supervision regime and with the baseline paper \cite{rost2026labelefficientunderwaterspeciesclassification}. The main low-label settings focus on small numbers of labeled examples per species, where label efficiency matters most.
Interesting to note is that due to class imbalance, some classes might receive more samples in the absolute experiment while some might receive more sampes in the fractional one. Future ablations might take the class support into account in order to generate a more consistent budgeting strategy, e.g by setting a minimum amount of labels per class.

Note that fractional budgets are applied to the post-validation training pool,
not to the full training split. Because 40\% of the training data is reserved
for circuit selection (Section~4.2), a fractional budget of, for example,
10\% corresponds to approximately 6\% of the original training data.
\subsection{Implementation and reporting}

All feature extraction is performed in PyTorch, and downstream classification is implemented in scikit-learn. Results are averaged across repeated runs with fixed random seeds for each circuit and classifier combination. The main reported metrics are macro F1-score and accuracy. In the class-specific experiment, we report the best-performing combination of circuit and downstream method for each class.

\section{Results}

\subsection{Circuit duplication improves frozen embeddings across all budgets}

Table~\ref{tab:global_results} presents the main global results. Across all seven label budgets, both the globally optimized circuit (Exp2) and the class-specific circuit selection (Exp3) improve macro F1 over the standard frozen baseline. The class-specific setting provides the largest gains: at 5 seeds per class, class-specific circuit selection reaches 0.735 macro F1 compared to 0.665 for the baseline, an improvement of 7.0 points. At 10 and 20 seeds per class, the gains are 6.6 and 4.9 points respectively. Even at the 100\% budget, class-specific selection improves over the baseline by 4.3 points.

The global circuit (Exp2) captures part of this improvement but not all. At every budget, the class-specific setting outperforms the global optimum, confirming that a single duplicated circuit cannot fully exploit the class-dependent structure in the data. Notably, the relative advantage of circuit duplication is largest at the lowest label budgets, which is exactly the regime where label efficiency matters most.

\begin{figure*}[t]
\centering
\includegraphics[width=\textwidth]{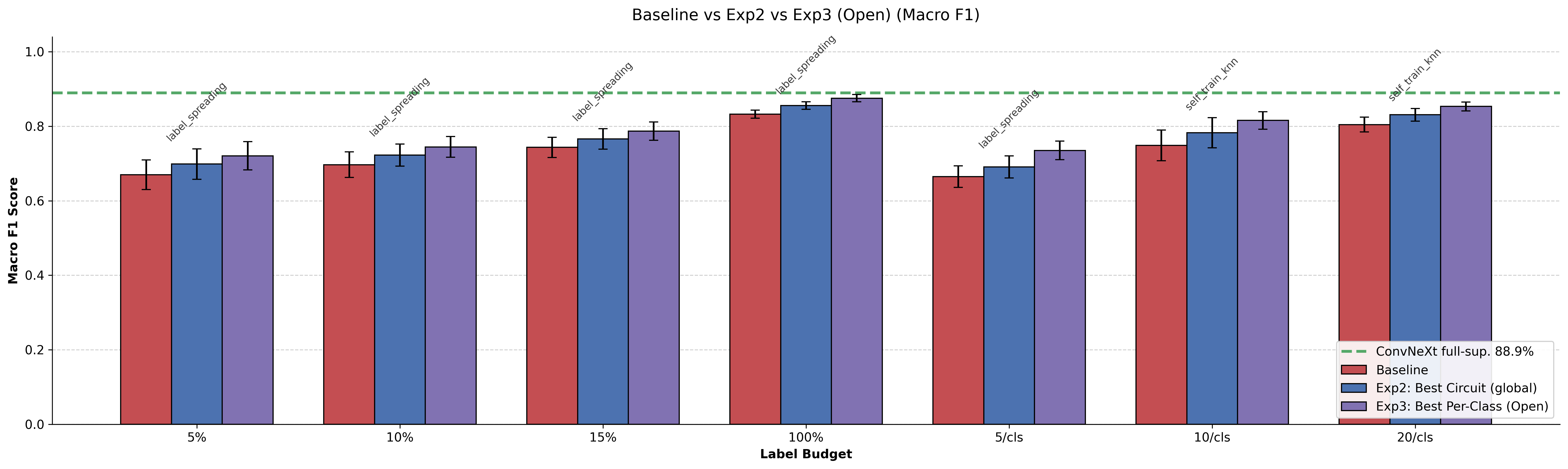}
\caption{Global macro F1 across label budgets. Red: standard frozen baseline. Blue: globally optimized circuit (Exp2). Purple: class-specific circuit selection (Exp3). The green dashed line marks the fully supervised ConvNeXt benchmark at 88.9\% \cite{fuad2026aqua20}. Circuit duplication improves over the baseline at every budget, with class-specific selection providing the largest gains. At 100\% labels, the gap to full supervision narrows to 1.4 percentage points.}
\label{fig:global_f1}
\end{figure*}

Figure~\ref{fig:global_f1} shows these results visually alongside the fully supervised ConvNeXt benchmark \cite{fuad2026aqua20}.

\begin{table*}[t]
\centering
\caption{Global test-set performance (Macro F1) across label budgets. The fully supervised ConvNeXt benchmark \cite{fuad2026aqua20} is shown for reference. Baseline refers to the standard frozen DINOv3 forward pass \cite{rost2026labelefficientunderwaterspeciesclassification}. Global Circuit selects a single best duplicated circuit for the full dataset. Class-Specific selects the best circuit independently per class. Bold indicates the best result among the frozen-embedding methods.}
\label{tab:global_results}
\begin{tabular}{@{}lcccc@{}}
\toprule
Budget & ConvNeXt & Baseline & Global Circuit & Class-Specific \\ \midrule
5 seeds/cls     & 0.889 & 0.665 & 0.691 & \textbf{0.735} \\
10 seeds/cls    & 0.889 & 0.749 & 0.782 & \textbf{0.815} \\
20 seeds/cls    & 0.889 & 0.804 & 0.830 & \textbf{0.853} \\
5\% seeds       & 0.889 & 0.670 & 0.698 & \textbf{0.720} \\
10\% seeds      & 0.889 & 0.697 & 0.723 & \textbf{0.744} \\
15\% seeds      & 0.889 & 0.743 & 0.766 & \textbf{0.787} \\
100\% seeds     & 0.889 & 0.832 & 0.855 & \textbf{0.875} \\
\bottomrule
\end{tabular}
\end{table*}

\subsection{Comparison to full supervision without training}

The comparison to the fully supervised ConvNeXt benchmark \cite{fuad2026aqua20} provides important context for these results. The ConvNeXt model was trained end-to-end on the full AQUA20 training set and achieves a macro F1 of 0.889. At the 100\% label budget, class-specific circuit selection reaches 0.875, closing the gap to just 1.4 percentage points. At 20 seeds per class, the gap is 3.6 points. These results are achieved entirely without gradient-based training: the DINOv3 backbone is frozen, no weights are updated, and the downstream classifiers are classical methods that require no backpropagation.

This finding is significant because it shows that a frozen foundation model, combined with inference-path optimization, can approach the performance of a fully supervised model trained specifically on the target dataset. The practical implication is that competitive classification performance may be achievable with only a fraction of the annotation effort and compute required by end-to-end training.

Moreover, as noted in Section~4.2, the downstream classifiers are trained on
only approximately 60\% of the original training data, since the remaining
40\% is reserved for circuit selection, making the comparison to ConvNeXt, which uses the full training split, conservative.

\subsection{Per-class analysis: circuit duplication exceeds full supervision for specific classes}

While the global macro F1 remains slightly below the fully supervised benchmark, the per-class analysis reveals a more nuanced picture. Table~\ref{tab:perclass_results} shows per-class F1 scores at the maximum label budget. Four classes exceed their ConvNeXt reference under class-specific circuit selection: octopus ($+$0.121), seaUrchin ($+$0.059), fishInGroups ($+$0.044), and starfish ($+$0.008).

The octopus result is particularly striking. The ConvNeXt model achieves an F1 of 0.750 for octopus, whereas class-specific circuit selection reaches 0.871, a gain of more than 12 points. The frozen baseline already surpasses ConvNeXt for this class (0.847), but circuit duplication extends the advantage further. This suggests that the DINOv3 representation already contains information about octopus that the supervised model fails to exploit, and that circuit duplication helps surface that information more effectively.

For the remaining classes, class-specific circuit selection narrows the gap to ConvNeXt in most cases. The average deficit across all 20 classes is only 1.4 F1 points ($\Delta = -0.014$), and for 14 out of 20 classes the gap is less than 4 points.

\begin{figure*}[t]
\centering
\includegraphics[width=\textwidth]{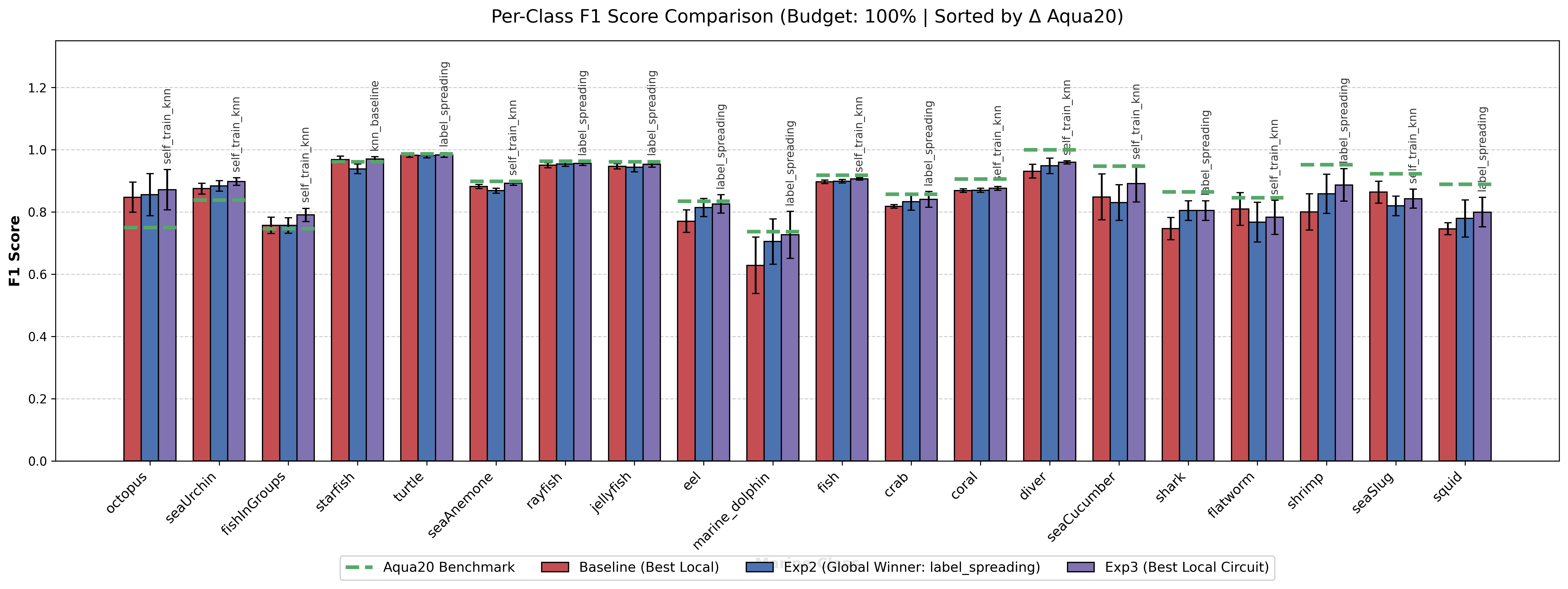}
\caption{Per-class F1 scores at the 100\% label budget. Red: best standard frozen baseline classifier per class. Blue: per-class score under the globally optimized circuit (Exp2). Purple: class-specific circuit selection (Exp3). Green dashed lines: ConvNeXt fully supervised benchmark per class \cite{fuad2026aqua20}. Classes are sorted by the difference between class-specific selection and the ConvNeXt reference. Octopus, seaUrchin, fishInGroups, and starfish exceed the fully supervised benchmark.}
\label{fig:perclass_100}
\end{figure*}

Figure~\ref{fig:perclass_100} shows the per-class comparison at the 100\% budget.

\begin{table*}[t]
\centering
\small
\caption{Per-class F1 score comparison at maximum label budget (100\% of the post-validation training pool; see Section~4.2). ConvNeXt: fully supervised benchmark \cite{fuad2026aqua20}. Baseline: best standard frozen DINOv3 classifier. Global Circuit: per-class score under the globally optimal circuit. Class-Specific: per-class best circuit and classifier. $\Delta$ is the difference between Class-Specific and ConvNeXt. Bold positive values indicate classes where the frozen-embedding approach exceeds full supervision.}
\label{tab:perclass_results}
\resizebox{\textwidth}{!}{%
\begin{tabular}{@{}llccccc@{}}
\toprule
Class & Best Method & ConvNeXt & Baseline & Global Circuit & Class-Specific & $\Delta$ \\ \midrule
coral           & self\_train\_knn   & 0.906 & 0.869 & 0.870 & 0.876 & $-$0.030 \\
crab            & label\_spreading   & 0.857 & 0.818 & 0.833 & 0.841 & $-$0.017 \\
diver           & self\_train\_knn   & 1.000 & 0.931 & 0.948 & 0.959 & $-$0.041 \\
eel             & label\_spreading   & 0.835 & 0.770 & 0.814 & 0.825 & $-$0.010 \\
fish            & self\_train\_knn   & 0.918 & 0.897 & 0.899 & 0.906 & $-$0.012 \\
fishInGroups    & self\_train\_knn   & 0.746 & 0.757 & 0.756 & 0.790 & \textbf{+0.044} \\
flatworm        & self\_train\_knn   & 0.846 & 0.810 & 0.767 & 0.810 & $-$0.037 \\
jellyfish       & label\_spreading   & 0.962 & 0.946 & 0.943 & 0.953 & $-$0.008 \\
marine\_dolphin & label\_spreading   & 0.737 & 0.628 & 0.705 & 0.726 & $-$0.011 \\
octopus         & self\_train\_knn   & 0.750 & 0.847 & 0.855 & \textbf{0.871} & \textbf{+0.121} \\
rayfish         & label\_spreading   & 0.963 & 0.950 & 0.954 & 0.956 & $-$0.007 \\
seaAnemone      & self\_train\_knn   & 0.899 & 0.882 & 0.868 & 0.892 & $-$0.006 \\
seaCucumber     & self\_train\_knn   & 0.947 & 0.848 & 0.830 & 0.892 & $-$0.056 \\
seaSlug         & knn\_baseline      & 0.923 & 0.864 & 0.819 & 0.864 & $-$0.059 \\
seaUrchin       & self\_train\_knn   & 0.839 & 0.875 & 0.883 & 0.898 & \textbf{+0.059} \\
shark           & label\_spreading   & 0.865 & 0.746 & 0.804 & 0.804 & $-$0.061 \\
shrimp          & label\_spreading   & 0.952 & 0.800 & 0.859 & 0.887 & $-$0.066 \\
squid           & label\_spreading   & 0.889 & 0.746 & 0.779 & 0.799 & $-$0.089 \\
starfish        & knn\_baseline      & 0.962 & 0.968 & 0.938 & 0.970 & \textbf{+0.008} \\
turtle          & label\_spreading   & 0.987 & 0.982 & 0.982 & 0.983 & $-$0.003 \\
\midrule
\textbf{Macro avg.}  &                    & 0.889 & 0.847 & 0.855 & 0.875 & $-$0.014 \\
\bottomrule
\end{tabular}
}
\end{table*}

\subsection{Class wide preference for specific circuits}

\begin{figure}[t]
\centering
\includegraphics[width=\columnwidth]{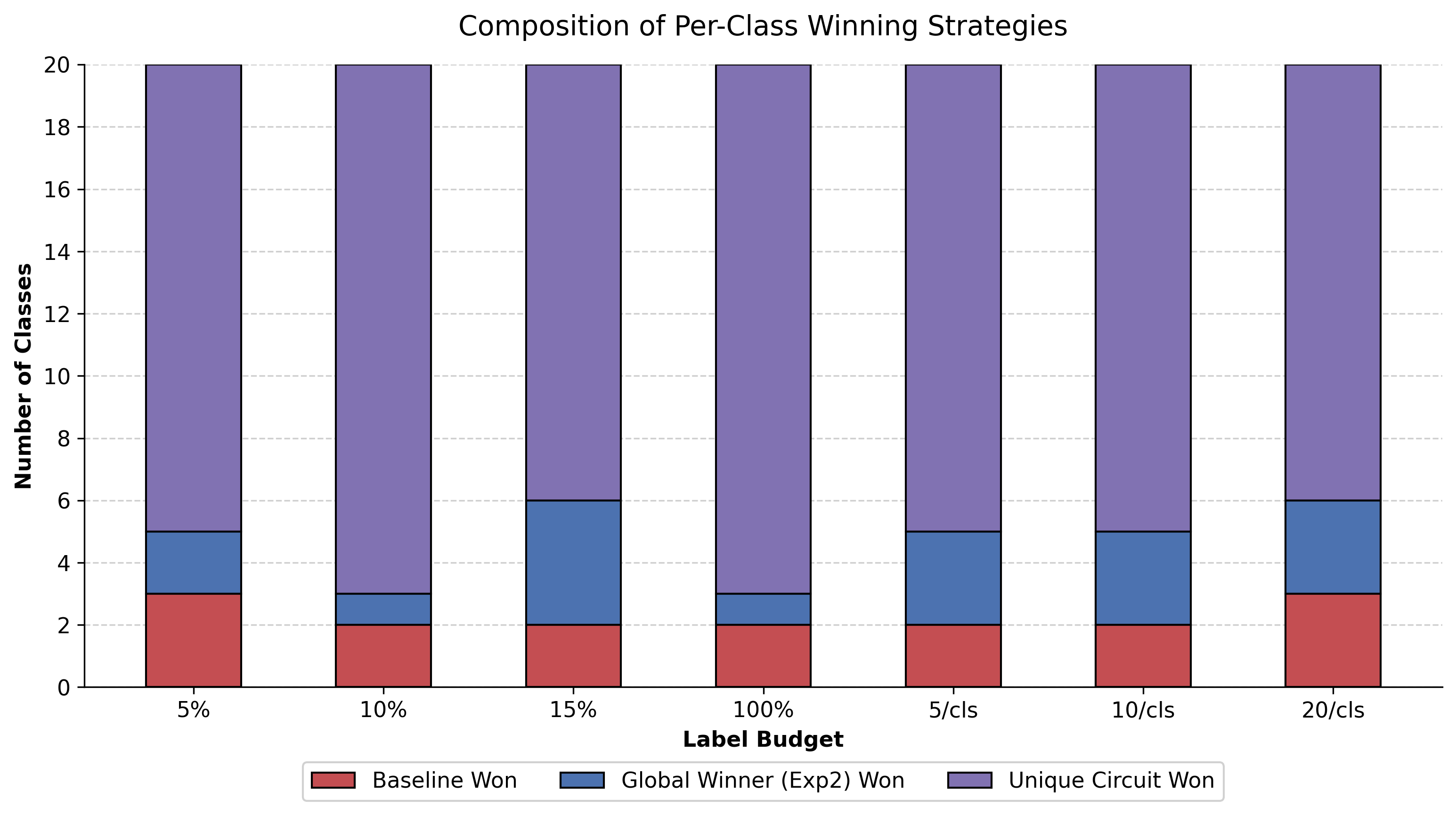}
\caption{Composition of per-class winning strategies across label budgets. Red: number of classes where the standard forward pass (baseline) achieves the best F1. Blue: classes where the globally optimized circuit is best. Purple: classes where a unique circuit, distinct from both baseline and global winner, achieves the best F1. Across all budgets, approximately 75\% of classes prefer a class-specific circuit.}
\label{fig:winning_strategy}
\end{figure}

Figure~\ref{fig:winning_strategy} shows, for each label budget, how many of the 20 classes are best served by the standard forward pass (baseline), by the globally optimal circuit, or by a unique circuit that differs from both. Across all seven budgets, only 2-3 classes prefer the standard forward pass, and approximately 15 classes prefer a circuit that is neither the baseline nor the global winner. This pattern is remarkably stable: it does not collapse at high label budgets.

This result has two implications. First, it confirms that circuit duplication is not simply compensating for label scarcity. Even when all training labels are available, different species benefit from different inference paths through the transformer. Second, it suggests that the visual properties relevant for distinguishing different marine species are distributed across different regions of the transformer's computational path, and that circuit duplication provides a way to selectively amplify those properties without any parameter modification.

\subsection{Downstream classifier patterns}

The per-class results in Table~\ref{tab:perclass_results} also reveal which downstream classifiers benefit most from circuit-modified embeddings. At the maximum budget, self-training KNN and label spreading are the dominant winning methods, accounting for all 20 classes between them. This indicates that the benefit of circuit duplication is not an artefact of one particular downstream classifier but is expressed through both graph-based semi-supervised methods and pseudo-labeling approaches.

\section{Discussion}

The results show that Circuit Duplication consistently improves downstream classification from frozen visual foundation model embeddings. Three findings deserve particular discussion.

\textbf{Approaching full supervision without training.} The most practically significant result is that class-specific circuit selection closes the gap to the fully supervised ConvNeXt benchmark to just 1.4 macro F1 points at the maximum label budget, and to 3.6 points with only 20 labeled examples per class. This is achieved entirely without gradient-based training. The DINOv3 backbone remains frozen, no parameters are updated, and the downstream classifiers are classical methods. This suggests that a substantial portion of the performance gap between frozen embeddings and end-to-end supervised training can be recovered by optimizing the inference path rather than the model weights.

\textbf{Exceeding full supervision for specific classes.} Four classes, octopus, seaUrchin, fishInGroups, and starfish, exceed their fully supervised ConvNeXt reference under class-specific circuit selection. The octopus result is especially notable: the ConvNeXt model achieves only 0.750 F1 for this class, likely because octopus is a rare class (10 test samples) with high visual variability and camouflage-like appearance. The frozen DINOv3 baseline already outperforms ConvNeXt at 0.847, and circuit duplication pushes this further to 0.871. This suggests that the general-purpose DINOv3 representation captures information about visually complex, low-frequency classes that a domain-specific supervised model fails to exploit, and that circuit duplication helps surface that information. More broadly, the fact that a training-free method can exceed full supervision for specific classes is an unusual result that warrants further investigation.

\textbf{Class-specific effects are stable and not driven by label scarcity.} The winning strategy distribution (Figure~\ref{fig:winning_strategy}) shows that approximately 75\% of classes prefer a unique circuit across all label budgets, including the full-label setting. This rules out the interpretation that circuit duplication is merely compensating for noisy estimates under extreme label poverty. Instead, it indicates that different species genuinely benefit from different computational paths through the transformer, likely reflecting differences in the visual properties (shape, texture, colour, context) that distinguish them. This observation connects to the broader question of whether different downstream tasks engage different parts of a foundation model's internal computation, and suggests that circuit-level optimization may be a useful complement to classifier-level optimization even when labels are plentiful.

\textbf{Headroom for improvement.} It is worth noting that the results reported here
were obtained under deliberately minimal downstream assumptions. No hyperparameter
tuning was performed on the downstream classifiers; all methods were run with default
or fixed settings. The classifiers themselves are among the simplest available:
nearest-neighbour methods, graph-based label propagation, and centroid-based
clustering. No ensemble methods, gradient-boosted models, or neural classifiers
were used. The foundation model was used entirely as provided, with no
domain-specific fine-tuning on underwater imagery, and no use of foundation models
natively trained on marine or ecological data. Each of these directions represents
a plausible route to further improvement. Taken together, the fact that a
macro F1 of 0.875 can be reached under such minimal conditions, within 1.4
points of a fully supervised, end-to-end trained ConvNeXt, suggests that
the effective ceiling of this approach, once these additional degrees of freedom
are explored, may well exceed the fully supervised baseline while remaining
far less expensive in terms of both annotation effort and compute.

In addition, the current evaluation design reserves 40\% of the training
split for validation, meaning the downstream classifiers never see
the full labeled pool; reducing this fraction or adopting more
sample-efficient circuit selection strategies could yield further gains.
\textbf{Practical implications.} In settings, such as marine science, where expert annotation is expensive and rare classes may have very few examples, these results suggest that useful performance gains are achievable without the cost of full supervised retraining. The method requires only forward passes through the existing model and simple downstream classifiers, making it computationally lightweight compared to fully supervised pipelines that require expensive equipment as well as technical expertise that can be hard to come by. This could support more efficient annotation strategies, quicker adaptation to new survey sites, and active-learning-style data collection pipelines.

\textbf{Limitations.} Several limitations should be noted. This study evaluates a single dataset (AQUA20), a single foundation model (DINOv3 Base), and extremely simple downstream classifiers. The method currently depends on a sweep over all 66 possible circuits, which, while cheap relative to gradient-based training, adds inference cost. 
In addition, the PCA dimensionality reduction is fitted transductively on the
combined train and test embeddings for each circuit configuration. This does not
expose test labels, and because PCA is an unsupervised method that captures
directions of maximal variance, the influence of test samples on the projection
axes is unlikely to introduce meaningful bias toward better classification.
Nonetheless, it is a departure from a strictly inductive setting, and the effect
of fitting PCA on training data alone should be verified in follow-up work. And finally, the mechanism by which circuit duplication improves performance is not yet understood: why duplicating specific transformer layers should improve embedding quality for specific classes remains an open question. In addition, the class-specific optimization setting requires per-class validation, which may not always be practical with very few labeled examples. Any broader generalisation of these findings to other datasets, models, or domains would require further empirical investigation.

\section{Conclusion}

This paper introduces Circuit Duplication to computer vision by applying it to a frozen DINOv3 backbone in a label-efficient marine species classification setting. To our knowledge, this is the first application of the method outside of language modelling and the first application in marine species classification.

The results demonstrate that inference-time modification of the computational path through a frozen visual transformer consistently improves downstream classification. Class-specific circuit selection reaches a macro F1 of 0.875 at the maximum label budget, closing the gap to the fully supervised ConvNeXt benchmark (0.889) to 1.4 points without any gradient-based training. Four species exceed their fully supervised reference, with the largest gain observed for octopus ($+$12.1 F1 points). Across all label budgets, approximately 75\% of classes prefer a class-specific circuit that differs from both the standard forward pass and the global optimum, indicating that the benefit of circuit duplication is genuinely class-dependent and not driven by label scarcity alone.

These findings suggest that frozen visual foundation model embeddings are not only useful as static feature extractors, but can be further optimized at the level of their computational path. Circuit duplication provides a lightweight, training-free mechanism for this optimization, with practical relevance for label-constrained scientific classification tasks. Future work should investigate the mechanism underlying circuit-dependent improvements, characterise which circuit regions benefit which types of visual classes, and test the approach on additional datasets, models, and domains.

\section*{Declaration of AI Use}

Generative Artificial Intelligence systems were used as writing and coding assistants during the preparation of this manuscript. All research concepts, experimental design, data analysis, and scientific interpretations are the sole work of the author.

\bibliographystyle{unsrt}
\bibliography{bib}

\clearpage
\appendix
\section{Additional Results}

\subsection{Global accuracy}

\begin{figure*}[h]
\centering
\includegraphics[width=\textwidth]{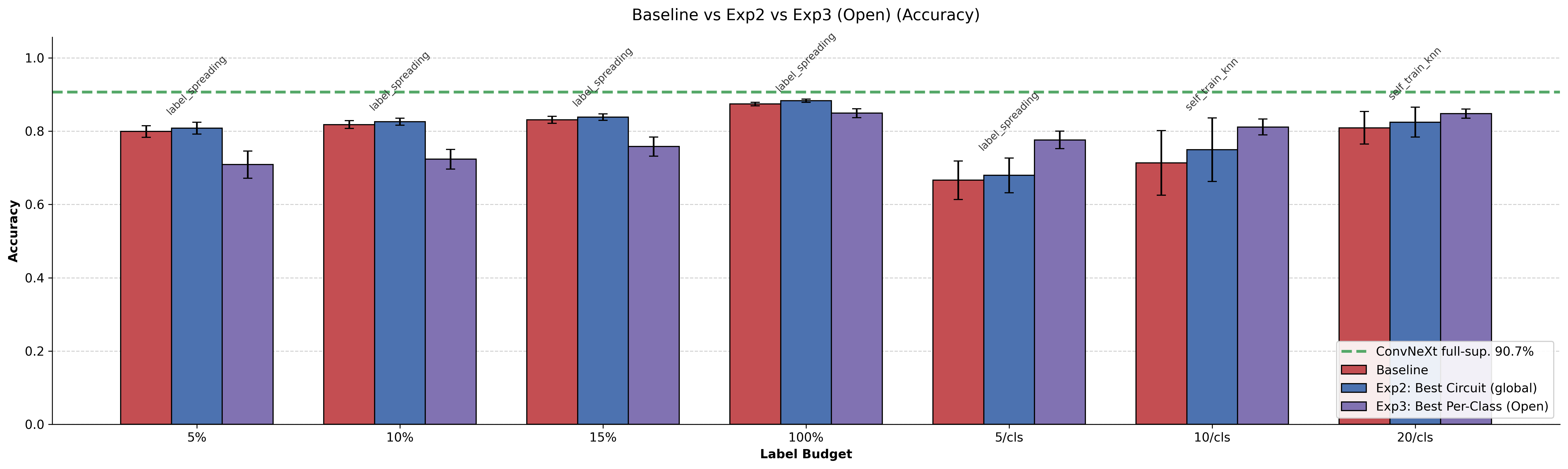}
\caption{Global accuracy across label budgets. Conventions as in Figure~\ref{fig:global_f1}. The ConvNeXt fully supervised benchmark is 90.7\%.}
\label{fig:global_acc}
\end{figure*}

Figure~\ref{fig:global_acc} shows the global accuracy comparison, complementing the macro F1 results in Figure~\ref{fig:global_f1}. The pattern is consistent: circuit duplication improves over the baseline at all budgets, with class-specific selection providing the largest gains.

\subsection{Per-class results across label budgets}

\begin{figure*}[h]
\centering
\includegraphics[width=\textwidth]{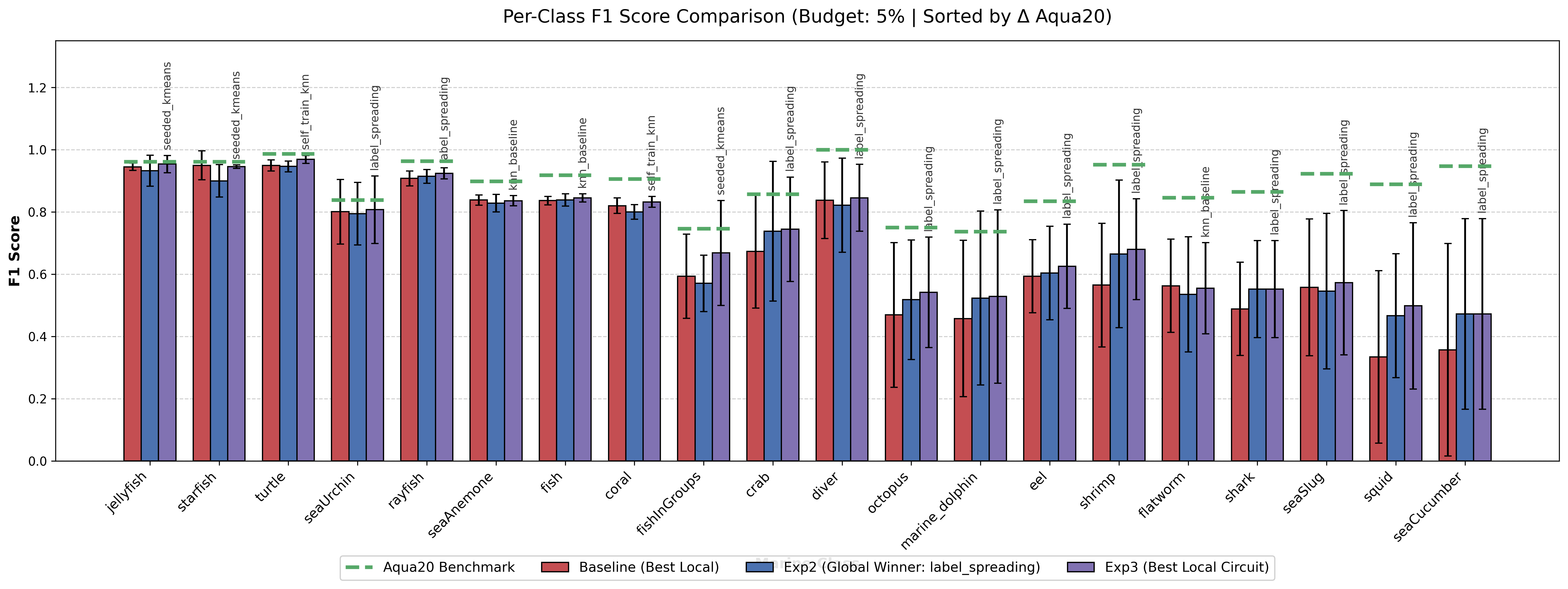}
\caption{Per-class F1 scores at 5\% label budget.}
\label{fig:perclass_5pct}
\end{figure*}

\begin{figure*}[h]
\centering
\includegraphics[width=\textwidth]{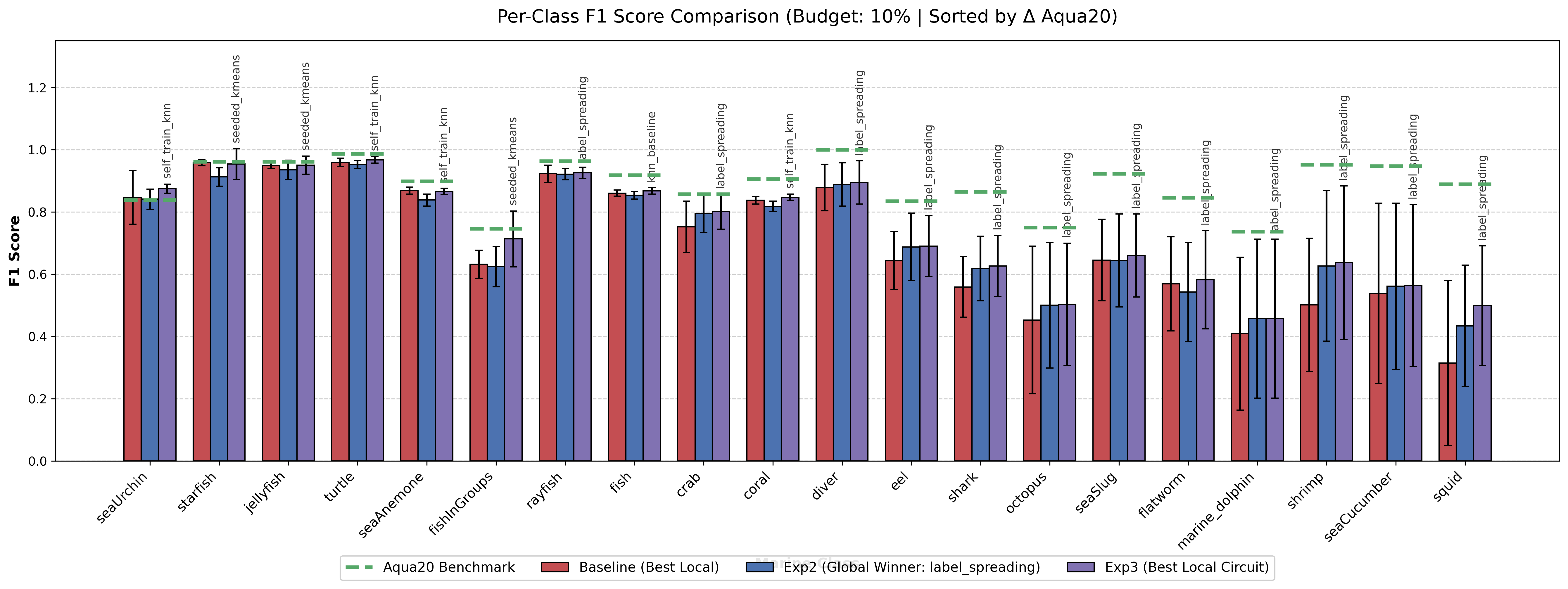}
\caption{Per-class F1 scores at 10\% label budget.}
\label{fig:perclass_10pct}
\end{figure*}

\begin{figure*}[h]
\centering
\includegraphics[width=\textwidth]{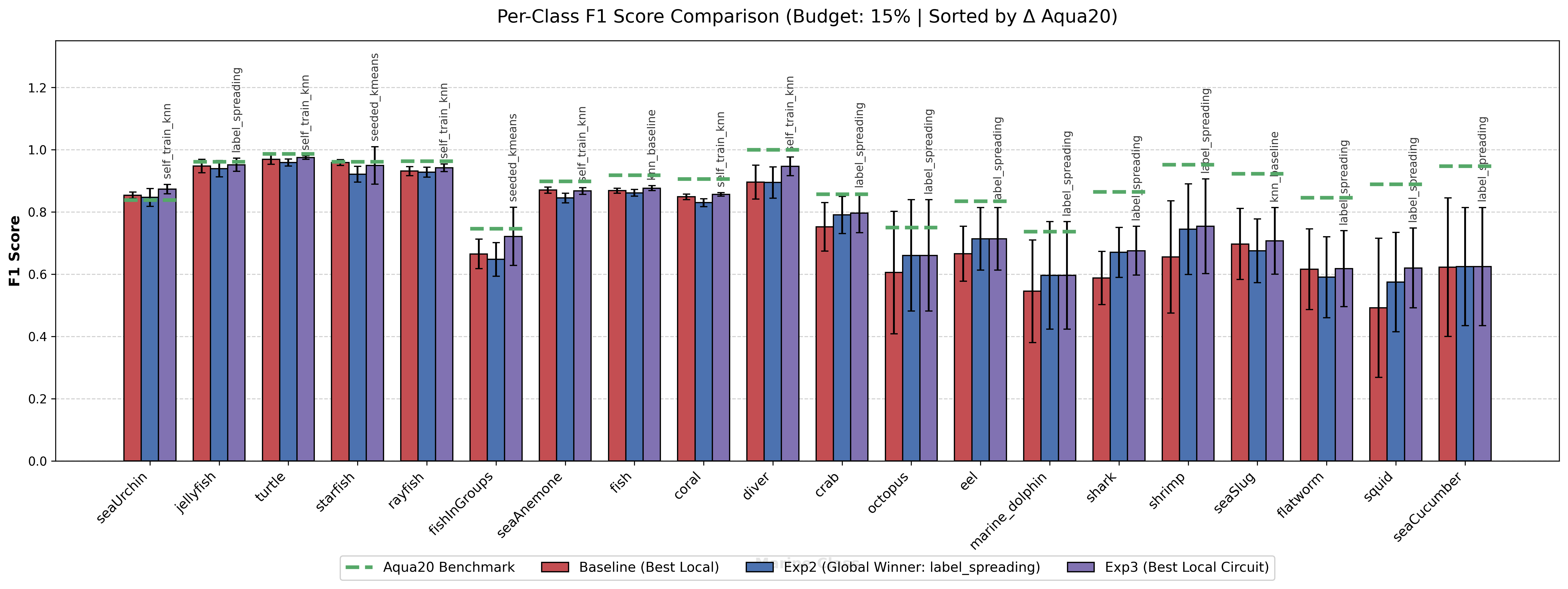}
\caption{Per-class F1 scores at 15\% label budget.}
\label{fig:perclass_15pct}
\end{figure*}

\begin{figure*}[h]
\centering
\includegraphics[width=\textwidth]{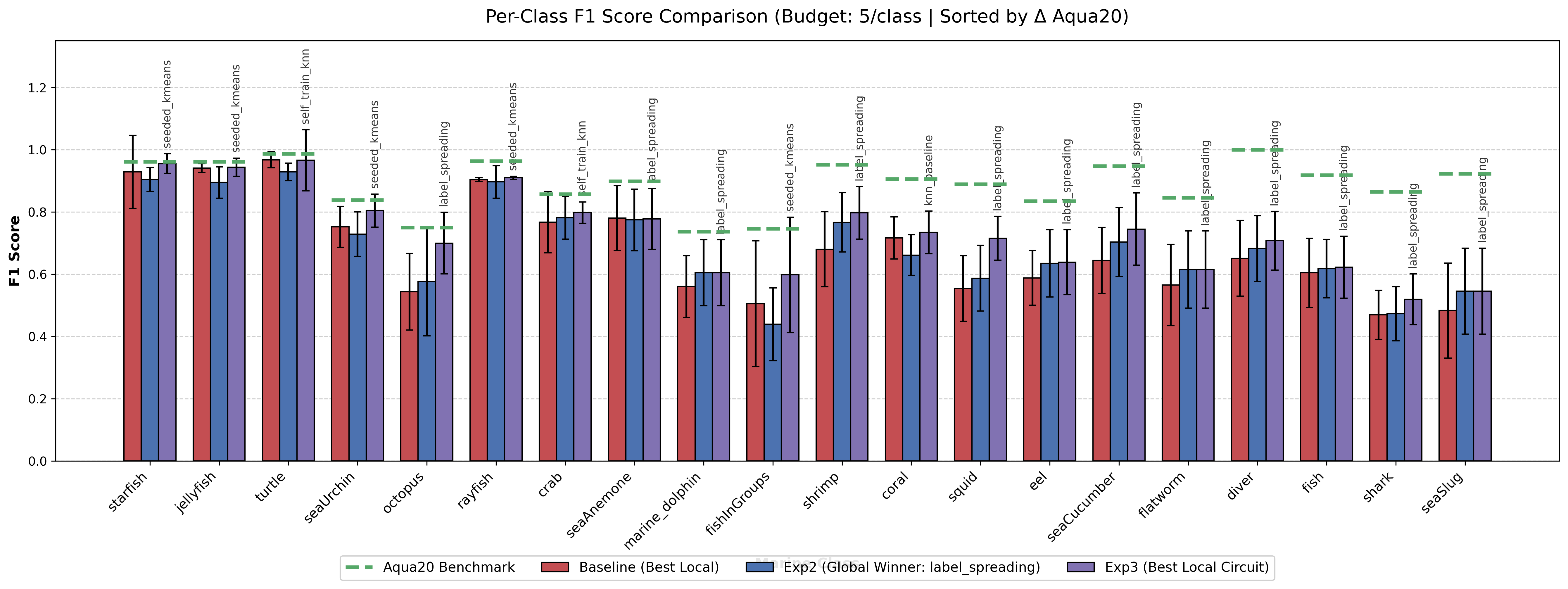}
\caption{Per-class F1 scores at 5 seeds per class.}
\label{fig:perclass_5cls}
\end{figure*}

\begin{figure*}[h]
\centering
\includegraphics[width=\textwidth]{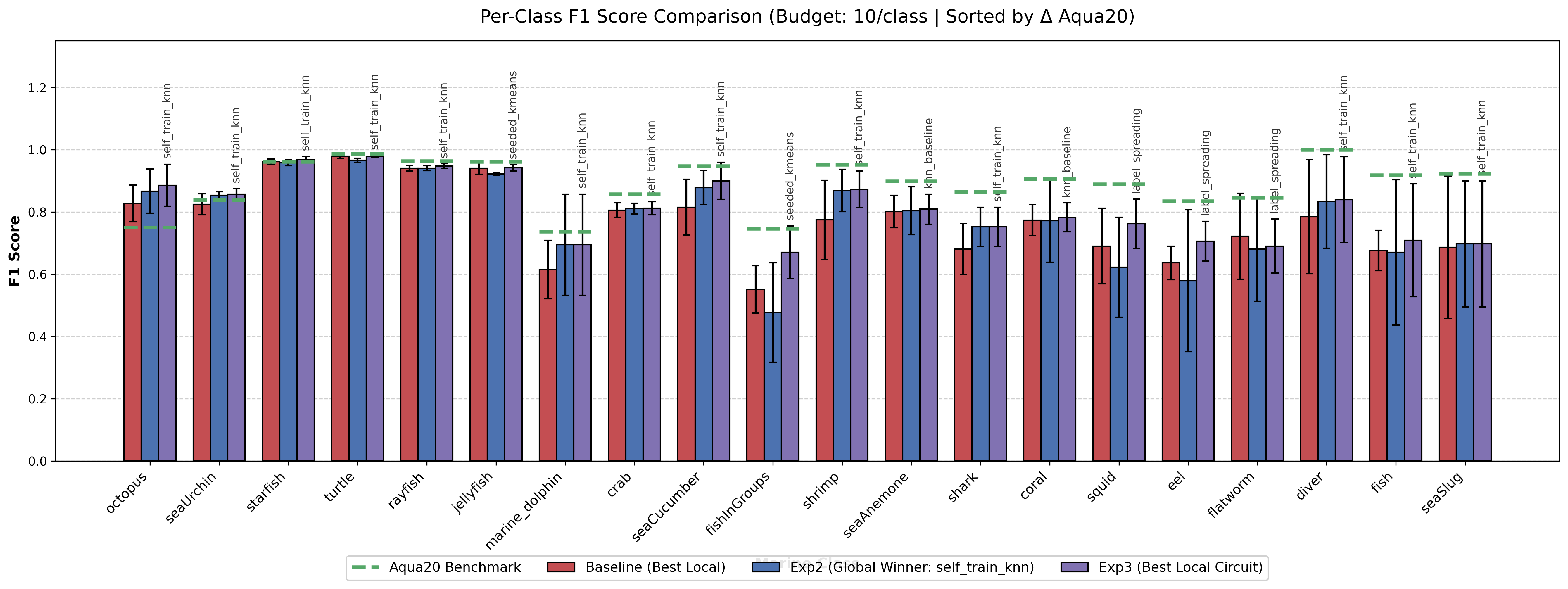}
\caption{Per-class F1 scores at 10 seeds per class.}
\label{fig:perclass_10cls}
\end{figure*}

\begin{figure*}[h]
\centering
\includegraphics[width=\textwidth]{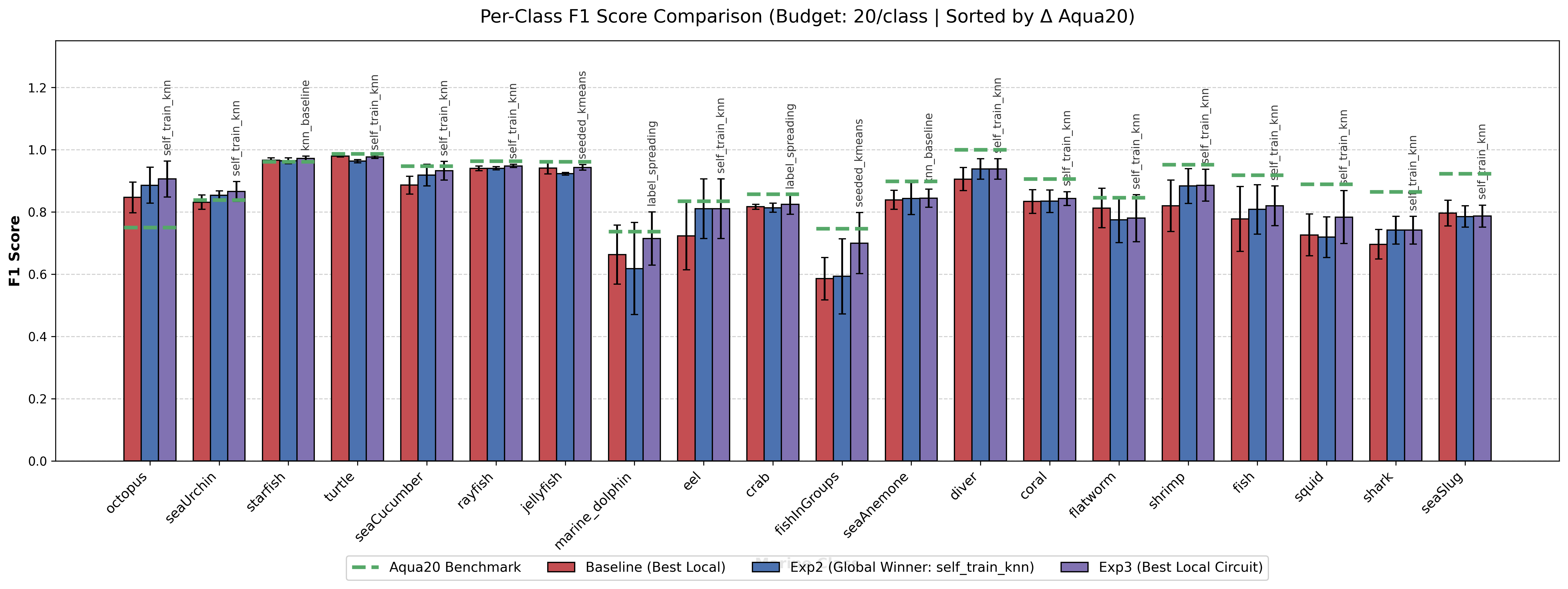}
\caption{Per-class F1 scores at 20 seeds per class.}
\label{fig:perclass_20cls}
\end{figure*}

Figures~\ref{fig:perclass_5pct}--\ref{fig:perclass_20cls} show per-class F1 scores at all evaluated label budgets. These complement Figure~\ref{fig:perclass_100} by showing how per-class patterns evolve from low- to high-label settings.

\end{document}

%% file: exp2_pipeline.tex
\begin{tikzpicture}[
    >=Stealth,
    node distance=0.5cm and 0.4cm,
    base/.style={draw, rounded corners=4pt, minimum height=1cm, font=\small, align=center, thick},
    gray box/.style={base, fill=gray!8, draw=gray!50, minimum width=2.2cm},
    teal box/.style={base, fill=teal!8, draw=teal!40, minimum width=2cm},
    purple box/.style={base, fill=violet!6, draw=violet!40},
    green box/.style={base, fill=green!6, draw=green!40},
    wide purple/.style={base, fill=violet!6, draw=violet!40, minimum width=12.8cm, minimum height=1.2cm},
    arr/.style={->, thick, gray!60},
    layer/.style={draw=gray!40, fill=gray!8, rounded corners=2pt, minimum width=0.5cm, minimum height=0.45cm, font=\tiny},
    hlayer/.style={draw=violet!60, fill=violet!15, rounded corners=2pt, minimum width=0.5cm, minimum height=0.45cm, font=\tiny},
]

\node[font=\large\bfseries] (title) {Global Circuit Selection};
\node[font=\small, gray!60, below=0.1cm of title] (sub) {Sweep 66 $(i,j)$ pairs, select one global best on val, report on test};

\node[gray box, below=0.6cm of sub.west, anchor=north west, xshift=1cm] (img) {\textbf{Image}\\[-2pt]{\scriptsize $518 \times 518$}};

\draw[arr] (img.south) -- ++(0, -0.5);

\node[purple box, minimum width=11cm, minimum height=1.8cm,
      below=0.5cm of img.south, anchor=north, xshift=2.8cm] (dinobox) {};
\node[font=\small\bfseries, violet!80!black] at ([yshift=0.55cm]dinobox.center) {Frozen DINOv3 ViT-B/16 with circuit duplication};

\node[layer] at ([xshift=-4.2cm, yshift=-0.15cm]dinobox.center) (l0) {\textsf{L0}};
\node[layer, right=0.08cm of l0] (ldot1) {\textsf{..}};
\node[hlayer, right=0.08cm of ldot1] (li1) {\textsf{Li}};
\node[hlayer, right=0.08cm of li1] (ldot2) {\textsf{..}};
\node[hlayer, right=0.08cm of ldot2] (lj1) {\textsf{Lj}};
\node[font=\tiny, gray!50, right=0.12cm of lj1] (sep) {|};
\node[hlayer, right=0.12cm of sep] (li2) {\textsf{Li}};
\node[hlayer, right=0.08cm of li2] (ldot3) {\textsf{..}};
\node[hlayer, right=0.08cm of ldot3] (lj2) {\textsf{Lj}};
\node[layer, right=0.08cm of lj2] (ldot4) {\textsf{..}};
\node[layer, right=0.08cm of ldot4] (ln) {\textsf{L11}};

\draw[decorate, decoration={brace, mirror, amplitude=4pt}, gray!50]
    ([yshift=-0.08cm]li1.south west) -- ([yshift=-0.08cm]lj1.south east)
    node[midway, below=4pt, font=\tiny, gray!60] {pass 1};
\draw[decorate, decoration={brace, mirror, amplitude=4pt}, gray!50]
    ([yshift=-0.08cm]li2.south west) -- ([yshift=-0.08cm]lj2.south east)
    node[midway, below=4pt, font=\tiny, gray!60] {pass 2};

\node[purple box, minimum width=1.6cm, right=0.5cm of dinobox.east, anchor=west] (emb) {\textbf{768-d}\\[-2pt]{\scriptsize $L_2$ norm}};
\draw[arr] (dinobox.east) -- (emb.west);

\node[teal box, minimum width=1.6cm, right=0.4cm of emb] (pca) {\textbf{PCA}\\[-2pt]{\scriptsize 128-d}};
\draw[arr] (emb) -- (pca);

\draw[arr] (pca.south) -- ++(0, -0.5);

\node[wide purple, below=0.5cm of pca.south, anchor=north east, xshift=0.4cm] (semi) {
    \textbf{Semi-supervised classification ($k$ labeled seeds per class)}\\[-2pt]
    {\scriptsize Label spreading \textbar\ Self-train KNN \textbar\ Self-train SVM \textbar\ Seeded K-means \textbar\ KNN baseline}
};

\coordinate (semi_drop) at ([xshift=-2.5cm]semi.south);
\draw[arr] (semi_drop) -- ++(0, -0.6);

\node[gray box, minimum width=5cm, below left=0.6cm and 0.5cm of semi.south, anchor=north] (sel) {\textbf{Select best global $(i,j)$}\\[-2pt]{\scriptsize on validation pool}};
\node[green box, minimum width=4.6cm, right=0.5cm of sel] (rep) {\textbf{Report on held-out test}};
\draw[arr] (sel) -- (rep);

\end{tikzpicture}

%% file: exp3_pipeline.tex
\begin{tikzpicture}[
    >=Stealth,
    node distance=0.5cm and 0.4cm,
    base/.style={draw, rounded corners=4pt, minimum height=1cm, font=\small, align=center, thick},
    gray box/.style={base, fill=gray!8, draw=gray!50, minimum width=2.2cm},
    teal box/.style={base, fill=teal!8, draw=teal!40, minimum width=2cm},
    purple box/.style={base, fill=violet!6, draw=violet!40},
    orange box/.style={base, fill=orange!8, draw=orange!50},
    green box/.style={base, fill=green!6, draw=green!40},
    wide purple/.style={base, fill=violet!6, draw=violet!40, minimum width=12.8cm, minimum height=1.2cm},
    arr/.style={->, thick, gray!60},
    layer/.style={draw=gray!40, fill=gray!8, rounded corners=2pt, minimum width=0.5cm, minimum height=0.45cm, font=\tiny},
    hlayer/.style={draw=violet!60, fill=violet!15, rounded corners=2pt, minimum width=0.5cm, minimum height=0.45cm, font=\tiny},
]

\node[font=\large\bfseries] (title) {Class-Specific Circuit Selection};
\node[font=\small, gray!60, below=0.1cm of title] (sub) {Each species gets its own optimal $(i,j)$ --- selected per class on val, report on test};

\node[gray box, below=0.6cm of sub.west, anchor=north west, xshift=1cm] (img) {\textbf{Image}\\[-2pt]{\scriptsize $518 \times 518$}};

\draw[arr] (img.south) -- ++(0, -0.5);

\node[purple box, minimum width=11cm, minimum height=1.8cm,
      below=0.5cm of img.south, anchor=north, xshift=2.8cm] (dinobox) {};
\node[font=\small\bfseries, violet!80!black] at ([yshift=0.55cm]dinobox.center) {Frozen DINOv3 ViT-B/16 with circuit duplication};

\node[layer] at ([xshift=-4.2cm, yshift=-0.15cm]dinobox.center) (l0) {\textsf{L0}};
\node[layer, right=0.08cm of l0] (ldot1) {\textsf{..}};
\node[hlayer, right=0.08cm of ldot1] (li1) {\textsf{Li}};
\node[hlayer, right=0.08cm of li1] (ldot2) {\textsf{..}};
\node[hlayer, right=0.08cm of ldot2] (lj1) {\textsf{Lj}};
\node[font=\tiny, gray!50, right=0.12cm of lj1] (sep) {|};
\node[hlayer, right=0.12cm of sep] (li2) {\textsf{Li}};
\node[hlayer, right=0.08cm of li2] (ldot3) {\textsf{..}};
\node[hlayer, right=0.08cm of ldot3] (lj2) {\textsf{Lj}};
\node[layer, right=0.08cm of lj2] (ldot4) {\textsf{..}};
\node[layer, right=0.08cm of ldot4] (ln) {\textsf{L11}};

\draw[decorate, decoration={brace, mirror, amplitude=4pt}, gray!50]
    ([yshift=-0.08cm]li1.south west) -- ([yshift=-0.08cm]lj1.south east)
    node[midway, below=4pt, font=\tiny, gray!60] {pass 1};
\draw[decorate, decoration={brace, mirror, amplitude=4pt}, gray!50]
    ([yshift=-0.08cm]li2.south west) -- ([yshift=-0.08cm]lj2.south east)
    node[midway, below=4pt, font=\tiny, gray!60] {pass 2};

\node[purple box, minimum width=1.6cm, right=0.5cm of dinobox.east, anchor=west] (emb) {\textbf{768-d}\\[-2pt]{\scriptsize $L_2$ norm}};
\draw[arr] (dinobox.east) -- (emb.west);

\node[teal box, minimum width=1.6cm, right=0.4cm of emb] (pca) {\textbf{PCA}\\[-2pt]{\scriptsize 128-d}};
\draw[arr] (emb) -- (pca);

\draw[arr] (pca.south) -- ++(0, -0.5);

\node[wide purple, below=0.5cm of pca.south, anchor=north east, xshift=0.4cm] (semi) {
    \textbf{Semi-supervised classification ($k$ labeled seeds per class)}\\[-2pt]
    {\scriptsize Label spreading \textbar\ Self-train KNN \textbar\ Self-train SVM \textbar\ Seeded K-means \textbar\ KNN baseline}
};

\coordinate (semi_drop) at ([xshift=-2.5cm]semi.south);
\draw[arr] (semi_drop) -- ++(0, -0.6);

\node[orange box, minimum width=5.6cm, below left=0.6cm and 0.3cm of semi.south, anchor=north] (sel) {\textbf{Select best $(i,j)$ per class}\\[-2pt]{\scriptsize on validation pool}};
\node[green box, minimum width=4.6cm, right=0.5cm of sel] (rep) {\textbf{Report on held-out test}};
\draw[arr] (sel) -- (rep);

\node[font=\scriptsize, orange!70!black, below=0.15cm of sel] {$\text{octopus} \to (i_1, j_1) \quad \text{jellyfish} \to (i_2, j_2) \quad \text{shark} \to (i_3, j_3) \quad \ldots$};

\end{tikzpicture}